\documentclass[10pt, final, journal, letterpaper, oneside, twocolumn]{IEEEtran}
\IEEEoverridecommandlockouts

\usepackage{graphicx} 

\usepackage{cite}
\usepackage{amsmath,amssymb,amsfonts}
\usepackage{algorithmic}
\usepackage{graphicx}
\usepackage{textcomp}
\usepackage{xcolor}
\usepackage{enumitem}
\usepackage{booktabs}
\usepackage{caption}
\usepackage{multirow}
\usepackage{makecell}
\usepackage{subcaption}
\usepackage[hidelinks]{hyperref}

\begin{document}

\title{Continual Learning for Adaptable Car-Following in Dynamic Traffic Environments}

\author{
    Xianda Chen,
    PakHin Tiu,
    Xu Han,
    Junjie Chen,
    Yuanfei Wu,
    Xinhu Zheng,
    Meixin Zhu
\thanks{\textit{Xianda Chen and PakHin Tiu are co-first authors.}}

\thanks{
    (\textit{Corresponding author: Meixin Zhu}) Xianda Chen, PakHin Tiu, Junjie Chen, Yuanfei Wu, Xinhu Zheng, and Meixin Zhu are with the Intelligent Transportation Thrust, The Hong Kong University of Science and Technology (Guangzhou), Guangzhou, 511400, China;
    Xu Han is with the Data Science and Analytics Thrust, Information Hub, The Hong Kong University of Science and Technology (Guangzhou), Guangzhou, China;
    Xinhu Zheng and Meixin Zhu are also with Guangdong Provincial Key Lab of Integrated Communication, Sensing and Computation for Ubiquitous Internet of Things.
    This study is supported by the National Natural Science Foundation of China under Grant 52302379 and 62373315, Guangzhou Basic and Applied Basic Research Projects under Grants 2023A03J0106, 2023A03J0683 and 2024A04J4290, Guangdong Province General Universities Youth Innovative Talents Project under Grant 2023KQNCX100, and Guangzhou Municipal Science and Technology Project 2023A03J0011.
    (email:
    xchen595@connect.hkust-gz.edu.cn,
    phtiu454@connect.hkust-gz.edu.cn,
    xhanab@connect.ust.hk,
    jchen321@connect.hkust-gz.edu.cn,
    yuanfeiwu@hkust-gz.edu.cn,
    xinhuzheng@hkust-gz.edu.cn,
    meixin@ust.hk
)
}
}

\maketitle

\begin{abstract}
    The continual evolution of autonomous driving technology requires car-following models that can adapt to diverse and dynamic traffic environments. Traditional learning-based models often suffer from performance degradation when encountering unseen traffic patterns due to a lack of continual learning capabilities. This paper proposes a novel car-following model based on continual learning that addresses this limitation. Our framework incorporates Elastic Weight Consolidation (EWC) and Memory Aware Synapses (MAS) techniques to mitigate catastrophic forgetting and enable the model to learn incrementally from new traffic data streams. We evaluate the performance of the proposed model on the Waymo and Lyft datasets which encompass various traffic scenarios. The results demonstrate that the continual learning techniques significantly outperform the baseline model, achieving 0\% collision rates across all traffic conditions. This research contributes to the advancement of autonomous driving technology by fostering the development of more robust and adaptable car-following models.
\end{abstract}

\section{Introduction}

\IEEEPARstart{T}{he} rapid development of autonomous driving technology promises to revolutionize transportation by enhancing both safety and efficiency. At the core of this technology lies the ability to navigate complex traffic scenarios, requiring models that can capture the dynamics of real-world traffic flow and make human-like decisions  \cite{zhu2018human, chen2024aggfollower, wei2021human,chen2024genfollower}. Car-following models play a crucial role in achieving this by simulating longitudinal control. These models enable autonomous vehicles to maintain safe distances from leading vehicles (LVs) in traffic streams \cite{chandler1958traffic, gazis1961nonlinear, pipes1953operational}. Adaptive Cruise Control (ACC) has been a valuable advancement in driver assistance, promoting safety and efficiency on highways by automating longitudinal control  \cite{xiao2010comprehensive, vahidi2003research}. However, traditional ACC systems often rely on pre-defined rules or control models. These approaches can be limited in dynamic traffic which is calibrated for typical car behavior, ACC might not adapt well to slower vehicles (trucks) or faster ones (motorcycles).

\begin{figure}[ht]
    \centering
    \includegraphics[width = 0.45\textwidth]{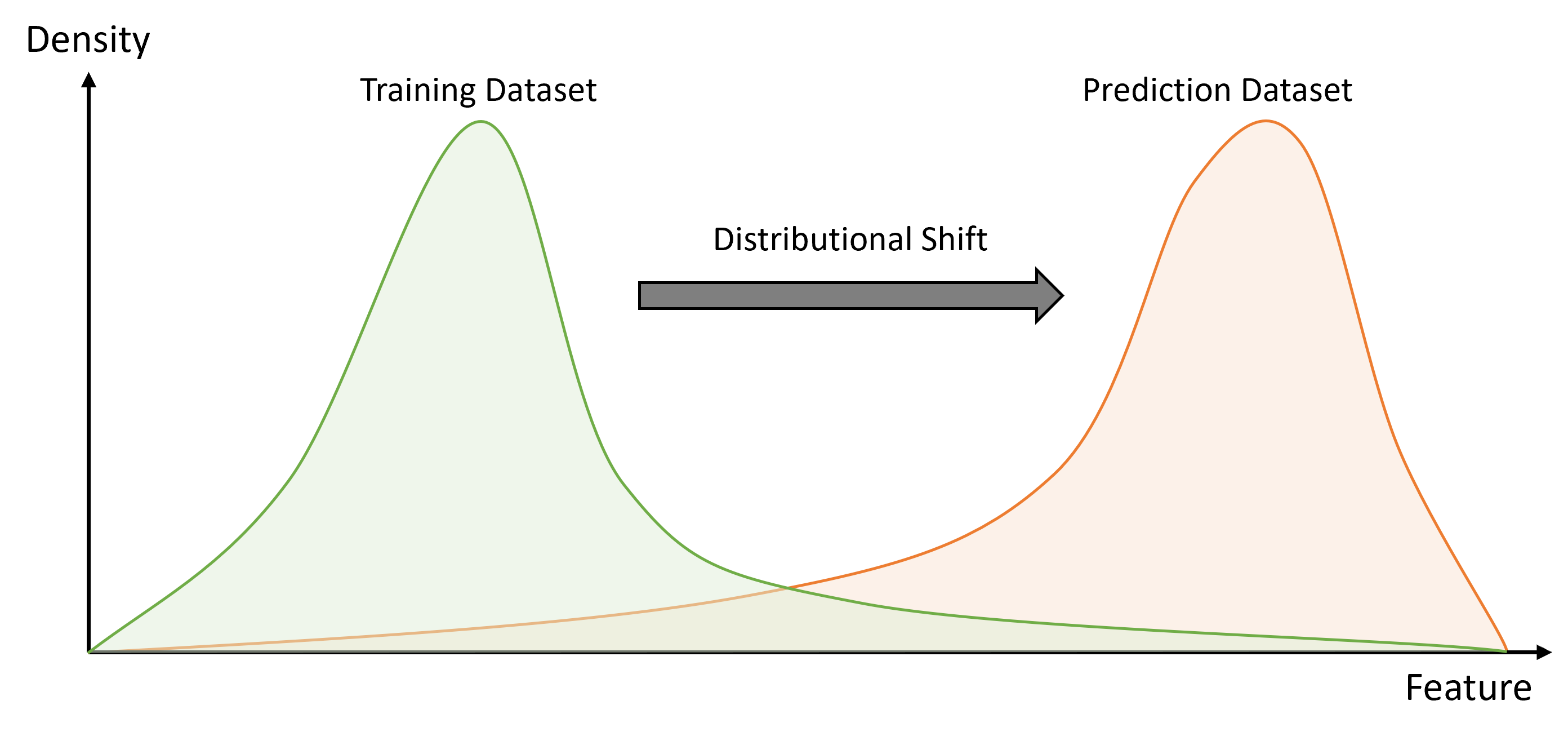}
    \caption{The densities of the training data (green) and prediction data (orange) reveal a distributional shift. This mismatch could lead to performance degradation on the prediction task.}
    \label{fig:distributional-shift}
\end{figure}

Learning-based methodologies, such as Q-learning, Deep Learning, and Reinforcement Learning, have emerged as promising solutions for car-following models \cite{zhu2020safe, hongfei2003develop, ma2020sequence, chen2024metafollower, chen2024editfollower}. However, these methods often rely heavily on the training data distribution (see Fig.~\ref{fig:distributional-shift}). This dependence can lead to performance degradation when encountering unforeseen situations that deviate from the training data. This vulnerability stems from the lack of continual learning capabilities, which refers to the ability to effectively adapt to new traffic patterns while retaining previously learned safe driving behaviors \cite{hadsell2020embracing, lesort2020continual}.

Retraining a car-following model from scratch for each novel scenario is not only computationally expensive and time-consuming but also risks catastrophic forgetting, where previously acquired knowledge is overwritten during the retraining process \cite{ramasesh2020anatomy, hayes2020remind}.

In light of these challenges, we propose a novel continual learning-based car-following model. By incorporating continual learning techniques, the proposed model aims to achieve robust performance by seamlessly adapting to new traffic situations without sacrificing previously learned safe driving behaviors. This approach fosters a more adaptable and generalizable car-following model, capable of thriving in the dynamic and diverse nature of real-world traffic environments.

This research contributes to the advancement of autonomous driving technology by introducing:
\begin{itemize}[leftmargin=*]
    \item We propose the implementation of continual learning approaches, specifically Elastic Weight Consolidation (EWC) and Memory Aware Synapses (MAS), for car-following models in autonomous vehicles. This enables the models to continuously learn and adapt to new traffic patterns encountered over time, unlike traditional approaches that struggle with new situations.
    \item We evaluate the performance of the proposed CL-based car-following models using extensive real-world datasets from Waymo and Lyft. This comprehensive evaluation demonstrates the models' effectiveness in handling unseen traffic scenarios, potentially leading to more reliable autonomous driving systems.
\end{itemize}

\section{Related Work}

Developing robust car-following models for autonomous vehicles is crucial. This section explores existing approaches and highlights the potential of continual learning.

\subsection{Car-Following Models}

Traditionally, car-following models fall into two categories: rule-based and data-driven (learning-based). Rule-based models offer efficiency and interpretability but struggle with dynamic traffic. Examples include Gipps' model (incorporates reaction time) \cite{gipps1981behavioural}, the GM model (considers speed and spacing) \cite{gazis1961nonlinear}, and the IDM (captures driver behavior) \cite{treiber2000congested}. Data-driven models learn from historical data to predict following vehicle (FV) behavior, offering flexibility for diverse conditions. Examples include: KNN~\cite{he2015simple}, Neural Networks (NN)~\cite{hongfei2003develop, yang2018novel, ma2020sequence}, and Reinforcement Learning (RL)~\cite{zhu2018human, zhu2020safe}. KNN models identify similar past situations but can suffer from the curse of dimensionality. NN-based models learn complex relationships from data but can be computationally expensive. RL-based models learn through trial and error in a simulated environment but can be complex to implement. However, data-driven models face a challenge: distributional shift. Real-world traffic can deviate from training data, leading to performance degradation in unseen scenarios. This necessitates continual learning.

\subsection{Continual Learning}

Continual learning (also lifelong learning) \cite{aljundi2018memory, kirkpatrick2017overcoming} allows models to adapt to new information continuously, overcoming limitations of static datasets. In computer vision, continual learning allows models to recognize new objects/scenes without forgetting old categories \cite{zhang2023slca}. Similarly, in natural language processing, it allows models to learn new vocabulary/writing styles while retaining proficiency in previously encountered patterns \cite{razdaibiedina2023progressive}. In autonomous driving, continual learning offers significant potential for car-following models. Traditional models struggle with unseen traffic scenarios. Continual learning equips them to learn from new data streams while preserving safe driving behaviors. Bao et al. \cite{bao2023lifelong} proposed a framework using generative replay to mitigate forgetting in vehicle trajectory prediction. Verwimp et al. \cite{verwimp2023clad} introduced CLAD, a continual learning benchmark for object classification/detection in autonomous driving. Continual learning holds promise for car-following models. By facilitating adaptation to evolving traffic patterns, it enhances model transferability and fosters safer, more reliable autonomous driving systems. Our research builds upon these advancements by presenting a novel continual learning framework for car-following models.

\section{Continual Learning Car-Following Model}

This section introduces our innovative approach to car-following model development, focusing on continual learning to overcome the shortcomings of traditional retraining methods. Instead of starting from scratch with each update, our framework adopts an incremental learning strategy.

\subsection{Baseline Model and Training}

The foundation of our framework lies in a well-established car-following model, a Long Short-Term Memory (LSTM) network \cite{ma2020sequence}. This baseline model is initially trained on a comprehensive traffic dataset encompassing diverse traffic scenarios. The training process aims to minimize a fundamental loss function, commonly the L2 loss. The L2 loss measures the difference between the model's predicted FV dynamic information and the actual behavior observed in the training data. 

In our baseline model settings, the loss function is defined as follows:
\begin{align}
    \mathcal{L}_{\text{Baseline}}(\theta) = \text{MSELoss}{(s, s^{*})} + \text{Collision Penalty}
    \label{eq:baseline-loss}
\end{align}
Here, $s$ represents the state of the simulated vehicle (SV), while $s^{*}$ denotes the state of the FV from the dataset. The collision penalty term accounts for the severity of potential collisions, encouraging the model to prioritize safe driving behaviors during training.

\subsection{Continual Learning Techniques}

Our framework incorporates two prominent techniques in the domain of continual learning for comparison: Elastic Weight Consolidation \cite{kirkpatrick2017overcoming} and Memory Aware Synapses \cite{aljundi2018memory}. These techniques address the challenge of catastrophic forgetting by prioritizing the preservation of previously learned knowledge during the learning of new traffic patterns.

\subsubsection{Elastic Weight Consolidation}

\begin{figure}[ht]
    \centering
    \includegraphics[width = 0.44\textwidth]{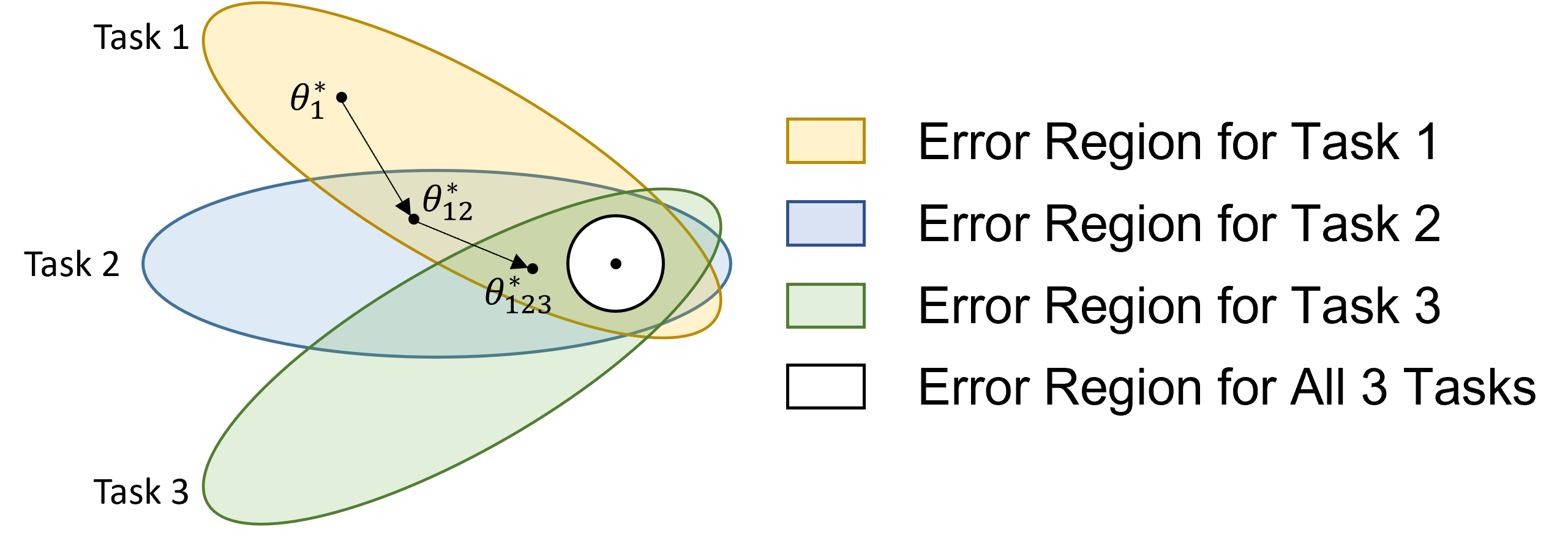}
    \caption{The impact of Elastic Weight Consolidation (EWC) on loss during continual learning. It illustrates how EWC helps mitigate the trade-off between learning new tasks and forgetting old tasks.}
    \label{fig:ewc}
\end{figure}

EWC incorporates a regularization term into the loss function to mitigate substantial alterations in the model's weights during the learning of new tasks. This additional term penalizes the model for making drastic weight updates that were pivotal for prior tasks, as illustrated in Fig.~\ref{fig:ewc}. By integrating a penalty factor based on the significance of each weight parameter, EWC ensures that previously acquired knowledge remains resilient to being overwritten.

Expressed mathematically, the EWC loss function takes the form:
\begin{align}
    \mathcal{L}_{\text{EWC}}(\theta) = \mathcal{L}_{\text{new}}(\theta) + \frac{\lambda}{2} \sum_{i}{F_{i}\left(\theta_{i} - \theta_{i}^{*}\right)^{2}}
    \label{eq:ewc-loss}
\end{align}
where $\mathcal{L}_{new}(\theta)$ denotes the base loss function for new tasks, $\lambda$ is a hyperparameter regulating the regularization term's influence, $F_{i}$ represents the Fisher information linked with the $i$-th weight parameter—an indicator of its importance, $\theta_{i}$ signifies the $i$-th weight parameter of the model, and $\theta_{i}^{*}$ represents the initial value of the $i$-th weight parameter prior to new task learning. Specifically, the diagonal entries of the Fisher information matrix at the old parameters $\theta_{i}^{*}$ are computed as:
\begin{align}
    F_{i} = \mathbb{E}_{x}{\left(\delta_{\theta_{i}} \log{p_{\theta}(x)}\right)^{2}}
    \label{eq:fisher-info}
\end{align}

Intuitively, the Fisher information gauges how the model's output varies with a slight adjustment in a particular weight. Weights possessing higher Fisher information values are deemed more crucial for past tasks and thus incur greater penalties from EWC for substantial updates during new task learning. This strategy encourages the model to prioritize adjusting less vital weights while preserving the foundational knowledge encoded in the more significant ones.

\subsubsection{Memory Aware Synapses}

MAS addresses catastrophic forgetting by dynamically adapting the learning rate of each weight parameter according to its significance in previously learned tasks. Weights identified as essential for past tasks are allocated lower learning rates during the training of new tasks. This tactic effectively decelerates their updates, reducing the risk of forgetting previously acquired safe driving behaviors.

Specifically, The MAS method introduces a weight importance modulation factor for each weight parameter. This factor dynamically adjusts the learning rate applied to the weight throughout the training process. Weights with greater importance for past tasks receive a lower modulation factor (closer to 0), leading to slower updates and a decreased likelihood of forgetting the knowledge stored in those weights.

The weight importance in MAS can be represented as:
\begin{align}
    \Omega_{i} = \frac{1}{N} \sum_{k=1}^{N}{\|g_{i}(x_{k})\|}
    \label{eq:mas-weight}
\end{align}
where $g_{i}(x_{k})$ denotes the gradient of the learned function concerning the parameter $\theta_{i}$ evaluated at the data point $x_{k}$, and $N$ represents the total number of data points in a given phase. After determining the weight importance $\Omega_{i}$, the MAS loss function is formulated as:
\begin{align}
    \mathcal{L}_{\text{MAS}}(\theta) = \mathcal{L}_{\text{new}}(\theta) + \lambda \sum_{i}{\Omega_{i}\left(\theta_{i} - \theta_{i}^{*}\right)^{2}}
    \label{eq:mas-loss}
\end{align}

Through the dynamic adjustment of learning rates based on weight importance, MAS ensures that the model concentrates on updating less critical weights during the learning of new tasks. This approach helps alleviate catastrophic forgetting and facilitates the continuous learning of new traffic patterns while safeguarding previously acquired safe driving behaviors.

\section{Experiments and Results}

\subsection{Dataset}

We've chosen the Waymo and Lyft datasets with $20,724$ car-following events based on criteria outlined in our previous work \cite{chen2023follownet}. These datasets offer rich sensor data derived from autonomous vehicles with diverse sensors and cameras, providing a comprehensive view of traffic environments; and diverse traffic scenarios encompassing urban and expressway scenarios, which is essential for training a robust model.

\begin{figure}
    \centering
    \includegraphics[width = 0.45\textwidth]{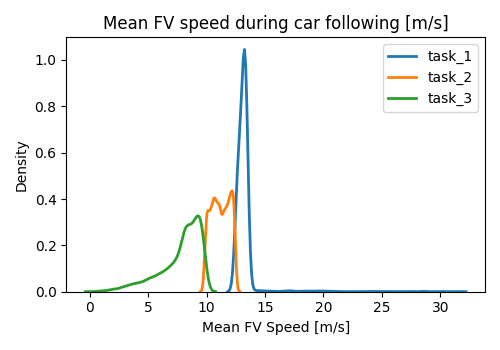}
    \caption{The distribution of mean FV speed across each event for all task sets.}
    \label{fig:task-distribution}
\end{figure}

To evaluate the continual learning capabilities of our model, we further divided the dataset into three distinct task sets based on different speed characteristics. This division is achieved by leveraging the percentiles of the mean speed of the FV across all events in the dataset, with the $33.3$rd and $66.7$th percentiles serving as splitting points. The $33.3$rd and $66.7$th percentiles of the mean FV speed are $9.86$ $m/s$ and $12.46$ $m/s$, respectively, resulting in three task sets with the following characteristics:
\begin{itemize}[leftmargin=*]
    \item \textbf{Task Set 1}: Mean FV speed range of $(12.46, \infty]$ $m/s$.
    \item \textbf{Task Set 2}: Mean FV speed range of $(9.86, 12.46]$ $m/s$.
    \item \textbf{Task Set 3}: Mean FV speed range of $[0, 9.86]$ $m/s$.
\end{itemize}
The distributions of all task sets are depicted in Fig.~\ref{fig:task-distribution}. This approach ensures each task set focuses on a distinct speed range, enabling assessment of the model's adaptability to new traffic patterns with differing speed characteristics. Finally, adhering to standard training practices, each task set is further divided into training ($70\%$), validation ($15\%$), and testing ($15\%$) sets.

\subsection{Evaluation Metrics}

This section details the metrics employed to evaluate the performance of the model compared to the baseline models.

\subsubsection{Metrics for Learning Performance}

We employ Mean Squared Errors (MSEs) in spacing and speed to quantify the difference between the predicted values of the FV generated by the model and the corresponding ground-truth values from the dataset.
Lower MSE values indicate better agreement between the model's predictions and the actual traffic behavior, reflecting the model's ability to learn effectively. The MSE for a set of predictions $y_{pred}$ and corresponding ground-truth values $y_{true}$ can be calculated using the following formula:
\begin{align}
    \text{MSE} = \frac{1}{N} \sum_{i}{(y_{pred, i} - y_{true, i})^{2}}
    \label{eq:mse}
\end{align}
where $N$ represents the total number of events in the testing dataset and $y_{pred, i}$ and $y_{true, i}$ represent the $i$-th predicted value and ground-truth value, respectively.

To evaluate the continual learning capabilities of our model, we calculate MSEs for all trained task sets with both the baseline model and the continual learning model at different stages of training:
\begin{itemize}[leftmargin=*]
    \item \textbf{Stage 1}: After training on Task Set 1 (initial learning)
    \item \textbf{Stage 2}: After training on Task Sets 1 and 2 (incremental learning)
    \item \textbf{Stage 3}: After training on all three task sets (final performance)
\end{itemize}
This approach allows us to assess how well the model learns from each task set and retains its knowledge when introduced to new traffic patterns.

\subsubsection{Metrics for Safety Performance}

We evaluate the safety performance of the car-following models by calculating the collision rate. This metric represents the frequency of collisions occurring during simulated test runs with the model controlling the SV. A lower collision rate indicates a safer car-following behavior by the model. Mathematically, the collision rate can be expressed as:
\begin{align}
    \text{Collision Rate} = \frac{\text{No. of Collisions}}{\text{Total No. of Events}} \times 100 \%
    \label{eq:collision-rate}
\end{align}

\subsection{Results}

\begin{figure*}
    \centering
    \includegraphics[width = 0.8\textwidth]{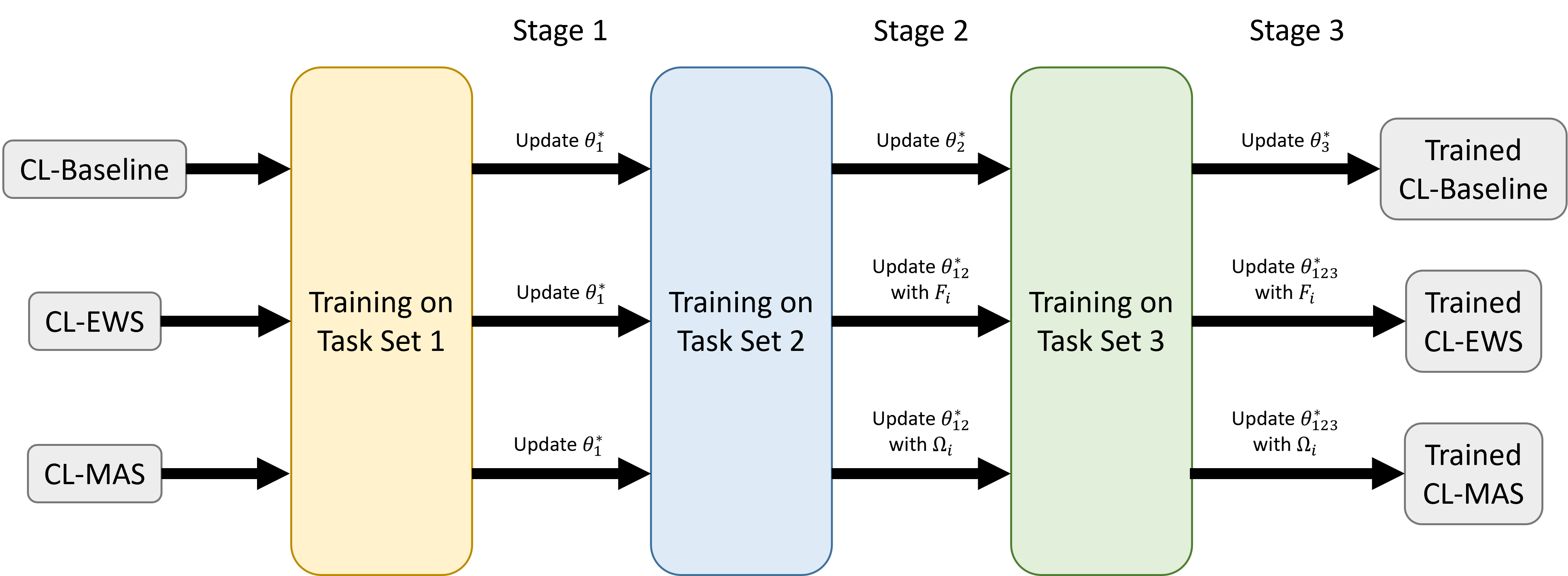}
    \caption{ Comparison of training flows for continual learning car-following models: Baseline vs. EWC vs. MAS.}
    \label{fig:training-flow}
\end{figure*}

This section presents the evaluation results of the proposed continual learning car-following models: continual learning with Elastic Weight Consolidation (CL-EWS) and continual learning with Memory Aware Synapses (CL-MAS). These models take the speed of the SV, the speed of the LV, the relative speed between them, and the spacing between them as inputs, and predict the appropriate acceleration value for the SV. We compare them to a continual learning baseline (CL-Baseline) and a standard LSTM model trained on all tasks simultaneously.

Figure~\ref{fig:training-flow} illustrates the training flow for each model. The CL-Baseline is trained incrementally on the three task sets one by one. However, it lacks any additional mechanism to address catastrophic forgetting. In contrast, CL-EWS and CL-MAS also undergo incremental training on the three tasks, but they incorporate an additional loss term during training. The specific loss terms used by CL-EWS and CL-MAS are detailed in Equations~\ref{eq:fisher-info} and~\ref{eq:mas-weight}, respectively.

\subsubsection{Training Hyperparameters}

For all LSTM models (regular LSTM, CL-Baseline, CL-EWS, and CL-MAS), we employed the following hyperparameters to ensure a consistent evaluation environment:
\begin{itemize}[leftmargin=*]
    \item Total epochs: $5$
    \item Historical horizon: $10$ (considering past $10$ timesteps)
    \item Learning rate: $0.001$
    \item Batch size: $32$
    \item Collision penalty and backward movement penalty: $1000$
    \item Random seed: $42$ 
\end{itemize}

\subsubsection{Performance on Individual Tasks}

\begin{table}[ht]
    \centering
    \begin{tabular}{c|c|ccc}
        \toprule
        Model & Task Set & Stage 1 & Stage 2 & Stage 3 \\
        \midrule
        \multirow{3}{*}{LSTM} & 1 & - & - &  22.64 \\
                              & 2 & - & - &  88.90 \\
                              & 3 & - & - & 136.77 \\
        \midrule
        \multirow{3}{*}{CL-Baseline} & 1 & 23.01 & 47.78 &  81.35 \\
                                     & 2 & -     & 59.91 & 117.46 \\
                                     & 3 & -     & -     & 112.70 \\
        \midrule
        \multirow{3}{*}{CL-EWS} & 1 & 3.01 & 2.75 & \textbf{8.02} \\
                                & 2 & -    & 3.18 & \textbf{3.64} \\
                                & 3 & -    & -    & \textbf{5.65} \\
        \midrule
        \multirow{3}{*}{CL-MAS} & 1 & 10.38 & 11.30 & 23.53 \\
                                & 2 & -     & 10.93 &  8.09 \\
                                & 3 & -     & -     &  7.48 \\
        \bottomrule
    \end{tabular}
    \caption{Performance comparison of MSE in spacing.}
    \label{tab:mse-spacing}
\end{table}

\begin{table}[ht]
    \centering
    \begin{tabular}{c|c|ccc}
        \toprule
        Model & Task Set & Stage 1 & Stage 2 & Stage 3 \\
        \midrule
        \multirow{3}{*}{LSTM} & 1 & - & - & 0.58 \\
                              & 2 & - & - & 2.48 \\
                              & 3 & - & - & 3.82 \\
        \midrule
        \multirow{3}{*}{CL-Baseline} & 1 & 0.59 & 1.22 & 2.33 \\
                                     & 2 & -    & 1.90 & 3.27 \\
                                     & 3 & -    & -    & 3.31 \\
        \midrule
        \multirow{3}{*}{CL-EWS} & 1 & 0.10 & 0.10 & \textbf{0.26} \\
                                & 2 & -    & 0.10 & \textbf{0.11} \\
                                & 3 & -    & -    & \textbf{0.23} \\
        \midrule
        \multirow{3}{*}{CL-MAS} & 1 & 0.33 & 0.35 & 0.67 \\
                                & 2 & -    & 0.36 & 0.26 \\
                                & 3 & -    & -    & 0.24 \\
        \bottomrule
    \end{tabular}
    \caption{Performance comparison of MSE in speed.}
    \label{tab:mse-speed}
\end{table}

Tables~\ref{tab:mse-spacing} and~\ref{tab:mse-speed} show the MSEs in spacing and speed for all models across the three tasks at different training stages.

The CL-Baseline suffers from catastrophic forgetting, with significant MSE increases in both spacing and speed after each new task. Compared to stage 1, the final MSEs increase by 253.5\% and 292.1\% for spacing and speed, respectively. This highlights the limitations of traditional continual learning approaches.

Both CL-EWS and CL-MAS models exhibit substantially lower MSE increases compared to the baseline. However, some interesting observations can be made regarding the differences between CL-EWS and CL-MAS. CL-EWS shows generally lower MSE in both spacing and speed across all stages, particularly for Task Set 1. This suggests that CL-EWS might be more effective in mitigating forgetting for entirely new tasks. On the other hand, CL-MAS performs better for Task Set 2, potentially indicating its ability to adapt to variations within previously encountered scenarios.

While both CL-EWS and CL-MAS achieve good overall performance, a more in-depth analysis is needed to fully understand the reasons behind the observed differences. Here are some potential factors to consider:
\begin{itemize}
    \item \textbf{Hyperparameter tuning}: The effectiveness of EWC and MAS can be sensitive to hyperparameter settings. It's possible that the chosen hyperparameters for CL-EWS were better suited for handling entirely new tasks, while CL-MAS benefited from parameters more appropriate for adapting to variations.
    \item \textbf{Model architecture}: The specific implementation details of CL-EWS and CL-MAS might influence their performance on different task types. Further investigation into the internal workings of each model could provide insights.
\end{itemize}

\begin{table}[ht]
    \centering
    \begin{tabular}{c|c|ccc}
        \toprule
        Model & Task Set & Stage 1 & Stage 2 & Stage 3 \\
        \midrule
        \multirow{3}{*}{LSTM} & 1 & - & - & \textbf{0.00} \\
                              & 2 & - & - & 1.03 \\
                              & 3 & - & - & 5.51 \\
        \midrule
        \multirow{3}{*}{CL-Baseline} & 1 & 0.19 & 0.00 & \textbf{0.00} \\
                                     & 2 & -    & 0.65 & 0.94 \\
                                     & 3 & -    & -    & 4.39 \\
        \midrule
        \multirow{3}{*}{CL-EWS} & 1 & 0.00 & 0.00 & \textbf{0.00} \\
                                & 2 & -    & 0.00 & \textbf{0.00} \\
                                & 3 & -    & -    & \textbf{0.00} \\
        \midrule
        \multirow{3}{*}{CL-MAS} & 1 & 0.00 & 0.00 & \textbf{0.00} \\
                                & 2 & -    & 0.00 & \textbf{0.00} \\
                                & 3 & -    & -    & \textbf{0.00} \\
        \bottomrule
    \end{tabular}
    \caption{Performance comparison of collision rate (\%).}
    \label{tab:collision-rate}
\end{table}

Table~\ref{tab:collision-rate} shows collision rates for all models across task sets and training stages. Notably, CL-EWS and CL-MAS maintain a perfect safety record (0\% collision rate) throughout training, unlike the baseline and standard LSTM models which exhibit higher collision rates. This highlights the safety benefits of continual learning with EWC or MAS.

\begin{figure*}[ht]
    \centering
    \begin{subfigure}[b]{0.32\textwidth}
        \includegraphics[width=\linewidth]{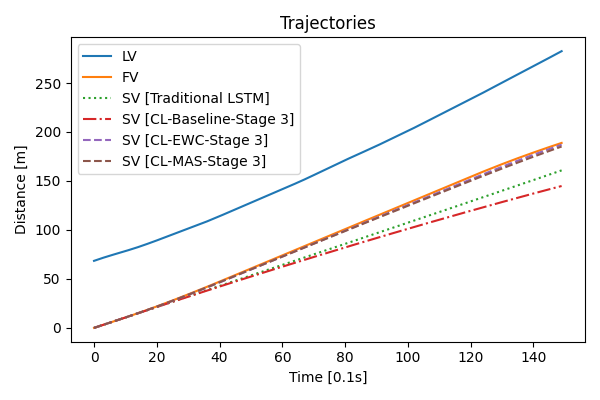}
        \caption{Random event selected from Task Set 1.}
        \label{fig:task-1-traj}
    \end{subfigure}
    \begin{subfigure}[b]{0.32\textwidth}
        \includegraphics[width=\linewidth]{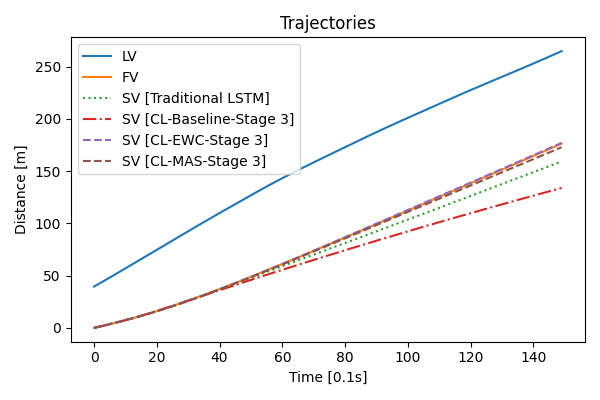}
        \caption{Random event selected from Task Set 2.}
        \label{fig:task-2-traj}
    \end{subfigure}
    \begin{subfigure}[b]{0.32\textwidth}
        \includegraphics[width=\linewidth]{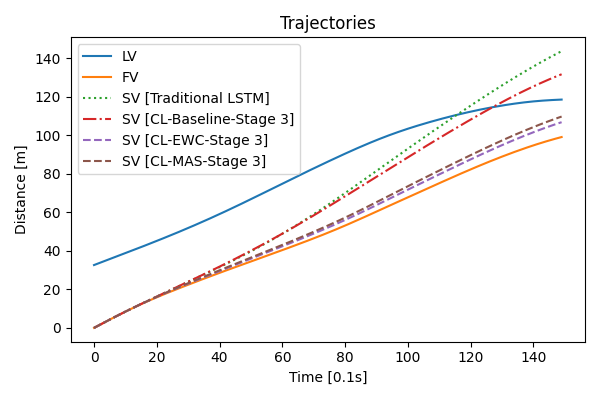}
        \caption{Random event selected from Task Set 3.}
        \label{fig:task-3-traj}
    \end{subfigure}
    \caption{Comparisons of trajectories performed by different models after training on all three task sets.}
    \label{fig:traj}
\end{figure*}

Fig.~\ref{fig:traj} compares trajectories of car-following models (LSTM: green dotted, CL-Baseline: red dashed-dotted, CL-EWS: purple dashed, CL-MAS: brown dashed) after training on all task sets. Notably, both CL-EWS and CL-MAS exhibit trajectories that closely follow the desired FV, unlike other models. This aligns with the prior results, suggesting that both CL-EWC and CL-MAS enable safe following behaviors.

\subsubsection{Key Observations}
\begin{itemize}
    \item \textbf{CL-EWS and CL-MAS maintain safe following distances}: As evident in Fig.~\ref{fig:traj}, the trajectories of CL-EWS (purple) and CL-MAS (brown) consistently maintain a safe distance behind the LV throughout the scenarios. This aligns with the 0\% collision rates observed in Table~\ref{tab:collision-rate}.
    \item \textbf{CL-EWS might be more cautious in unfamiliar situations}: While both models achieve safe following, CL-EWS trajectories appear slightly more conservative, particularly in areas with sudden changes (e.g., lane merging in Fig.~\ref{fig:task-3-traj}). This reinforces the idea that CL-EWS prioritizes retaining knowledge of entirely new tasks.
    \item \textbf{CL-MAS demonstrates adaptability within safe margins}: CL-MAS trajectories (brown) show a slightly more dynamic behavior compared to CL-EWS, particularly in Task Set 2 scenarios (Fig.~\ref{fig:task-2-traj}). This suggests CL-MAS's ability to adapt to variations in following distances while maintaining safety.
\end{itemize}

These observations are consistent with the findings from the MSE analysis. CL-EWS might be more conservative in its learning approach, prioritizing safety in entirely new situations. CL-MAS demonstrates a stronger ability to adapt to variations within similar scenarios while also achieving safe following behavior. This suggests a potential benefit of CL-MAS for handling continuously evolving traffic conditions, as it can adapt to new situations while maintaining performance on previously learned ones.

While both CL-EWS and CL-MAS exhibit significant safety improvements, it's important to acknowledge limitations. The current analysis focuses on simulated scenarios. Real-world traffic conditions can be far more complex and unpredictable. Future work can involve testing these models in more comprehensive driving simulation environments or potentially even on real-world datasets (if available while ensuring privacy concerns are addressed). Additionally, research can explore techniques to further enhance the safety and adaptability of these continual learning models for car-following tasks in autonomous driving systems.

\section{Conclusion}

This work addressed the challenge of catastrophic forgetting in car-following models by proposing a novel continual learning framework. Our approach leverages EWC and MAS to enable continual learning from new traffic data while retaining previously acquired knowledge of safe driving behaviors. Evaluations on a comprehensive dataset demonstrate that the proposed CL-EWS and CL-MAS models significantly outperform the baseline, achieving lower errors and maintaining a perfect safety record (0\% collision rate) across all traffic scenarios. These results highlight the potential of continual learning techniques in developing robust and adaptable car-following models, paving the way for safer and more reliable autonomous driving systems. Future research directions include exploring additional continual learning methods and integrating this framework with other components of an autonomous driving system.

\bibliographystyle{IEEEtran}
\bibliography{ref}

\end{document}